\documentclass[11pt,a4paper]{article}
\usepackage[hyperref]{acl2022}
\usepackage{times}
\usepackage{textcomp}
\usepackage{latexsym}
\usepackage{multicol}

\usepackage{xcolor}
\usepackage{multirow}
\usepackage{graphicx}
\usepackage{amsfonts}
\usepackage{amsmath}
\usepackage{algorithm}
\usepackage{booktabs}
\usepackage{algorithm}
\usepackage{soul} 
\usepackage[noend]{algpseudocode}

\DeclareMathOperator*{\argmin}{arg\,min}

\usepackage{microtype}

\aclfinalcopy 

\title{Unsupervised Text Deidentification}

\author{
    John X. Morris \ \ \ \  Justin T. Chiu\  \ \ \  Ramin Zabih\ \ \ \  Alexander M. Rush\\
    Cornell University \\
    \texttt{\{jxm3\}@cornell.edu} 
}

\date{}

\begin{document}
\maketitle
\begin{abstract}
 Deidentification seeks to anonymize textual data prior to distribution. Automatic deidentification primarily uses supervised named entity recognition from human-labeled data points. We propose an unsupervised deidentification method that masks words that leak personally-identifying information. The approach utilizes a specially trained reidentification model to identify individuals from redacted personal documents. Motivated by K-anonymity based privacy, we generate redactions that ensure a minimum reidentification rank for the correct profile of the document. To evaluate this approach, we consider the task of deidentifying Wikipedia Biographies, and evaluate using an adversarial reidentification metric. Compared to a set of unsupervised baselines, our approach deidentifies documents more completely while removing fewer words. Qualitatively, we see that the approach eliminates many identifying aspects that would fall outside of the common named entity based approach. \footnote{Our code and deidentified datasets are available \href{http://github.com/jxmorris12/unsupervised-text-deidentification}{on Github.}}

\end{abstract}

\section{Introduction}

In domains such as law, medicine, and government, it can be difficult to release textual data because it contains sensitive personal information~\cite{Johnson2016Mimic,Jana2021DiffPriv,Pilan2022TabAnonymizationBenchmark}. Privacy laws and regulations vary by domain and impact the  requirements for deidentification. Most prior work on automatic deidentification \cite{Neamatullah2008AutomatedDO,Meystre2010DeidSurvey,Sanchez2014UtilPresPriv,Liu2017DeidRnnCrf,Norgeot2020Philter,Sberbank2020DeidEmnlp} deidentifies data to the requirements of the HIPAA Safe Harbor method \cite{HIPAA1996}. Annotations for these systems are based on a list of 18 identifiers like age, phone number, and zip code. These systems treat deidentification as a named entity recognition problem within this space. Upon the removal of these pre-defined entities, text is no longer considered sensitive.

However, one of the 18 categories defined by HIPAA Safe Harbor includes ``any unique identifying number, characteristic, or code [that could be used to reidentify an individual]''. Prior work ignores this nebulous 18th category. One reason the category is ill-defined is due to the existence of \textit{quasi-identifiers}, pieces of personally identifiable information (PII) that do not fall under any single category and therefore are difficult to identify and label in the general case \cite{Phillips2016Discombob}. Even data that has all of the categories from Safe Harbor removed may still be reidentified through quasi-identifiers \cite{Angiuli2015HowTo}.  Supervised approaches cannot naturally detect quasi-identifiers, since these words are not inherently labeled as PII \cite{Uzuner2007EvaluatingDeid}.

\begin{figure*}[t]
    \hspace*{-0.2cm}\includegraphics[width=1.05\textwidth]{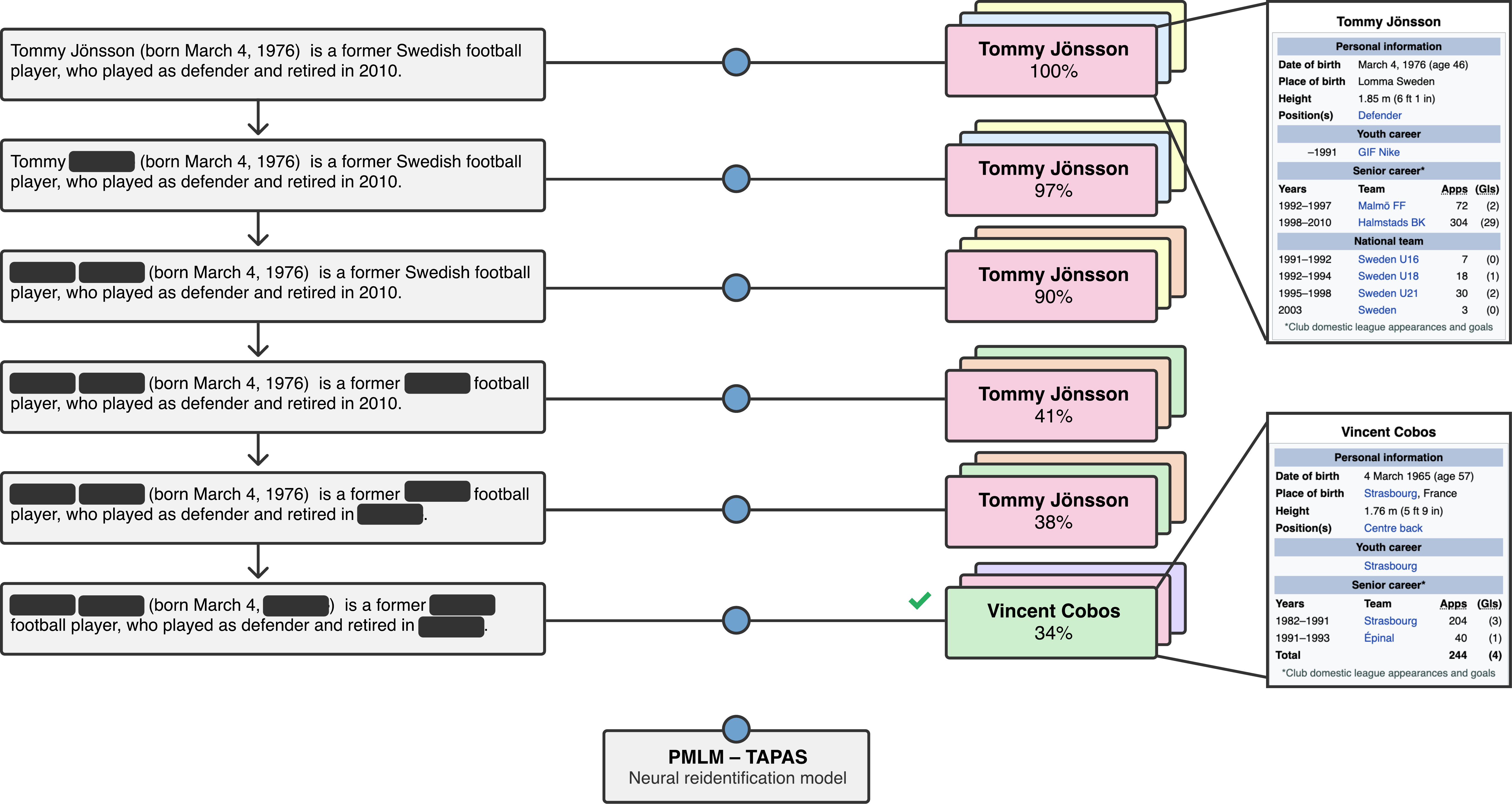}
    \caption{Method overview. A document ($x$, top-left) paired with a profile ($\hat{y}$,  top-right) is given to the system. 
        A trained neural reidentification model ($p(y | x, z)$, blue circle) produces a distribution over all possible profiles based on densely encoded representations. At each stage of inference, masks are added to the source document, changing the relative rank of the reidentification model. The method is run until k-anonymity of the reidentification model is achieved. Note that in this example, it is not necessary to remove all information, such as the month and day of birth, since the player is already deidentified.}
    \label{fig:fig1}
\end{figure*}

In this work, we propose an unsupervised deidentification method that targets the more general definition of PII. Instead of relying on specific rule lists of named entities, we directly remove words that could lead to reidentification. Motivated by the goal of \textit{K-anonymity} \cite{LisonAnonymizationAcl2021}, our approach utilizes a learned probabilistic reidentification model to predict the true identity of a given text. We perform combinatorial inference in this model to find a set of words that, when masked, achieve K-anonymity. The system does not require any annotations of specific PII, but instead learns from a dataset of aligned descriptive text and profile information. Using this information, we can train an identification process using a dense encoder model. 

Experiments test the ability of the system to deidentify documents from a large-scale database. We use a dataset of Wikipedia Biographies aligned with info-boxes \cite{LebretEMNLP2016Wikibio}. The system is fit on a subset of the data and then asked to deidentify unseen individuals. Results show that even when all words from the profile are masked, the system is able to reidentify $32\%$ of individuals. When we use our system to deidentify documents, it is able to fully anonymize them while retaining over 50\% of words. When we compare our deidentification method to a set of unsupervised baselines, our method deidentifies documents more completely while removing fewer words. We qualitatively and quantitatively analyze the redactions produced by our system, including examples of successfully redacted quasi-identifiers.

\section{Related Work}

\paragraph{Automated deidentification.} There is much prior work on deidentifying text datasets, both with rule-based systems \cite{Neamatullah2008AutomatedDO, Meystre2010DeidSurvey, Sanchez2014UtilPresPriv, Norgeot2020Philter, Sberbank2020DeidEmnlp} and deep learning methods \cite{Liu2017DeidRnnCrf, YuePHICON2020, Johnson2020BertDeid}. Each of these methods is supervised, relies on datasets with human-labeled PII, and focuses on removing some subset of the 18 identifying categories from HIPAA Safe Harbor. Other approaches include generating entire new fake datasets using Generative Adversarial Networks (GANs) \cite{ChinCheong2019GANsEHR}. \citet{Friedrich2019AdversarialNlpEmr} train an LSTM on an EMR-based NLP task using adversarial loss to prevent the model from learning to reconstruct the input. Finally, differential privacy is a technique for ensuring provably private distributions~\cite{Dwork2006DP}. It has mostly been used for training anonymized models on data containing PII, but requires access to the un-anonymized datasets for training \cite{Li2021DPLiang}. Our deidentification approach does not provide the formal guarantees of differential privacy, but aims to provide a practical solution for anonymizing datasets in real-world scenarios.


\paragraph{Deidentification by reidentification.} The NeurIPS 2020 Hide-and-Seek Privacy Challenge benchmarked both deidentification and reidentification techniques for clinical time series data \cite{Jordon2021NeurIPSChallenge}. In computer vision, researchers have proposed learning to mask faces in images to preserve the privacy of individuals using reidentification \cite{Hukkelas2019DeepPrivacyGan,Maximov2020CIAGAN,Gupta2021AdvMaskGen}. In NLP, some work has been done on evaluating the reidentification risk of deidentified text \cite{Scaiano2016ReidRisk}. \citet{ElEmam2009Kanon} proposes a method for deidentification of tabular datasets based on the concept of K-anonymity. \citet{Gardner2009Integrated} deidentify unstructured text by performing named entity extraction and redacting entities until k-anonymity. \citet{Mansour2021QuasiCloud} propose an algorithm for deidentification of tabular datasets by quantifying reidentification risk using a metric related to K-anonymity. In our work, we train a reidentification model in an adversarial setting and use the model to deidentify documents directly. 

\paragraph{Learning in the presence of masks.} Various works have shown how to improve NLP models by masking some of the input during training. \citet{Chen2020VMASK} show that learning in the presence of masks can improve classifier interpretability and accuracy. \citet{Li2016RepErasure} train a model to search for the minimum subset of words required that, when removed, change the output of a classifier. They apply their method to neural network interpretability, and use reinforcement learning. \citet{Liao2020Pmlm} pre-train a BERT-style language model to do masked-word prediction by sampling a masking ratio from $U(0,1)$ and masking that many words. While their method was originally proposed for text generation, we apply the same masking approach to train language models for redaction.


\section{Motivating Experiment: Quasi-Identifiers}
\label{sec:motivate}

In order to study the problem of deidentifying personal information from documents, we set up a model dataset utilizing personal profiles from Wikipedia. We use the Wikibio dataset \cite{LebretEMNLP2016Wikibio}. Each entry in the dataset contains a \textit{document}, the introductory text of the Wikipedia article, and a \textit{profile}, the infobox of key-value pairs containing personal information. We train on the training dataset of $582,659$ documents and profiles. During test time, we evaluate only test documents, but consider all $728,321$ profiles from the concatenation of the train, validation, and test sets. This dataset represents a natural baseline by providing a range of factual profile information for a large collection of individuals, making it challenging to deidentify. In addition, it provides an openly available collection for comparing models.

Is it difficult to deidentify individuals in this dataset? Wikipedia presents no domain challenges, and so finding entities is trivial.
In addition many of the terms in the documents overlap directly with the terms in the profile table. Simple techniques should provide robust deidentification. 

\begin{table}[t]
\centering
\begin{tabular}{lccc}
\toprule
& & \textbf{DeID} & \\
& \textbf{None} & \textbf{Named entity}  & \textbf{Lexical} \\
\textbf{ReID}  & (0\%) & (24\%) & (28\%)  \\
\midrule
\begin{tabular}[c]{@{}l@{}}IR ReID \end{tabular} & 74.9 & 4.3 & 0.0  \\ 
\begin{tabular}[c]{@{}l@{}}NN ReID\end{tabular} & 99.6 &  79.7  & 31.9 \\ 
\bottomrule
\end{tabular}
\caption{Percentage of documents reidentified (ReID) for different deidentification methods. Percentage of words masked in parentheses.}
\label{table:reid}
\end{table}

We test this with two deidentification techniques: (1) \textit{Named entity} removes all words in documents that are tagged as named entities. (2) \textit{Lexical} removes all words in the document that also overlap with the profile. To reidentify, we use an information retrieval model (BM25) and a dense neural network approach (described in Section~\ref{sec:model}). 

Table~\ref{table:reid} shows the results. While IR-based ReID is able to reidentify most of the original documents, without named entities or lexical matches, documents appear to be no longer reidentifiable. However, our model is able to reidentify 80\% of documents, even with all entities removed. With all lexical matches with the profile removed (32\% of total words), NN ReID is still able to reidentify a non-trivial number of documents. 

This experiment indicates that even in the WikiBio domain, there are a significant number of pseudo-identifiers that allow the system to identify documents \textit{even} when almost all known matching information is removed. 
In this work we study methods for discovering and quantifying these identifiers.

\section{Deidentification by Inference}

An overview of our data and system is shown in Figure~\ref{fig:fig1}. Given a document $x_1 \ldots x_N$, we consider the problem of uniquely identifying the corresponding person $y$ from a set of possible options $\cal Y$. 
The system works in the presence of redactions defined by a latent binary mask $z_1 \ldots z_N$ on each position, where setting $z_n=1$ masks word $x_n$. 

We define a reidentification model as a model of $p(y\ |\ x, z)$ that gives a probability to each profile in $\cal Y$ for a masked document. During deidentification, we assume that we have access to the true identity $\hat{y}$ of the document that we would like to hide.

Our objective is to find the minimally sized mask that ensures that $\hat{y}$ is not in the top-$K$ predictions of the identification model:
\begin{eqnarray*} 
\min_{z_1 \ldots z_N}  && |z| \\
\text{ s.t. } && \hat{y} \not \in K\arg\max_{y} p(y\ |\ x, z).
\end{eqnarray*}
This objective is motivated by the concept of $K$-Anonymity \citep{K-anonymity}. A dataset has K-anonymity if each person $\hat{y}$ in the dataset is indistinguishable from at least $K$ other people in $\mathcal{Y}$.

\begin{algorithm}[t]
\caption{Greedy Deidentification}
\label{alg:fbs}
\begin{algorithmic}
\State $x, \hat{y} \gets \mathrm{input \ document\ and\  person}$
\State $z_j \gets 0$ for all $j$
\For{$i = 1$  to $N$}
    \State{$j^* \gets \argmin_j p(y=\hat{y}\ |\ x, z_{-j}, z_j = 1)$}
    \State{$z_{j^*} \gets 1$}
    \If {$\hat{y} \not \in K\arg\max_{y} p(y\ |\ x, z)$} 
    \State \Return $z$
    \EndIf
\EndFor
\end{algorithmic}
\end{algorithm}

The $K$-anonymity objective is combinatorial, and is intractable to solve with a non-trivial reidentification model. We instead approximate it with search. Specifically we use a simple greedy deidentification technique shown in Algorithm~\ref{alg:fbs}.

\section{Reidentification Model}
\label{sec:model}

The core of this redaction system is a model of reidentification, $p(y \ | \ x, z)$.
Defining this model faces two challenges: a) to facilitate informed search in the presence of masks and b)
to correctly identify a person from 100,000s of choices. 

As we do not have access to supervised masks, 
we define the probability of unmasked identification as marginalizing over all possible masks:

\[ p(y\ |\ x) =  \mathbb{E}_{z\sim p(z\ |\ x)} p(y\ |\ x, z; \theta) \]

\noindent where $p(z \ | \ x)$ is the mask prior and $p(y\ |\ x, z; \theta)$ is the reidentification model.

To assign a prior over masks $p(z \ | \ x)$, we opt for a simple setting that avoids building in additional information and fits well with deidentification search. One possibility would be to follow BERT-style masking and mask words at a fixed ratio of 15\% ~\citep{DBLP:conf/naacl/DevlinCLT19}. However, \citet{Liao2020Pmlm} argue that while successful for classification, fixed-ratio masking works poorly for generation-style objectives. Following this advice, we use the following algorithm to construct masks of varying size: 

\begin{itemize}
    \item Sample the number of masks $l \sim \text{Uni}(0, N)$.
    \item Sample $l$ masked words $z_m$ by uniformly sampling indices $m$ from $\{1,\ldots, M\}$ without replacement.
\end{itemize}

For the reidentification model, $p(y | x, z; \theta)$, we follow the dense retrieval literature and use an embedding-based model~\citep{Karpukhin2020DPR}. Specifically we use an (asymmetric) bi-encoder model on documents and profiles. 
The document encoder $f$ computes an embedding of the masked document, and the profile encoder $g(y)$ produces an embedding of the profile table corresponding to person $y$. 
We score the match by computing the joint encoding $f(x, z)^{\top} g(y)$ using the dot product between the vectors outputted by two neural networks. 
Define the matrix of profile embeddings as $\mathbf{G} = [g(y_1); ...; g(y_{|\cal Y|})]$.   The reidentification probability is defined as
\[p(y=i\ |\ x, z) = \text{softmax}(f(x, z)^{\top} \mathbf{G})_i.\]
During training we utilize label smoothing on the distribution, which has also been shown to be useful when training for inference in an argmax setting~\cite{muller2019does}.

To train the model we optimize a lower bound on the identification log-likelihood: 
\[  \log p(y | x)  \geq   \mathbb{E}_{z\sim p(z | x)}[ \log p(y | x, z)] \] 

\noindent Specifically we sample a word dropout mask $z$ for each element $x$ from the prior, and then mask words during reidentification training.  

Note that for training we compute the full distribution and do not use a contrastive approximation. In order to learn the parameters of $g$ we utilize coordinate ascent. Specifically we fix $\mathbf{G}$ and optimize the parameters of $f$. We then switch and optimize the profile encoder $g$ on odd-numbered epochs to predict documents in $\cal X$ (with no masking), and then recompute $\mathbf{G}$. 

\begin{table*}[t]
\centering
\begin{tabular}{lccc}
\toprule
 & Privacy & \multicolumn{2}{c}{Utility} \\ 
\textbf{Method} & \textbf{Ensemble ReID} & \textbf{\% Masked} & \textbf{Info. Loss (\%)} \\ 
\midrule

(No redaction) & 99.6 & 0.0 & 0.0 \\
Lexical redaction & 31.9 & 32.1 & 20.8  \\ 
Named entity redaction & 79.7 & 27.3 & 27.3 \\

\midrule
\multicolumn{4}{c}{\textbf{$\mathbf{< 25\%}$ Reidentifiable}} \\ 
IDF & 24.2 & 58.5 & 66.3 \\
IDF (Table-Aware) & 21.2 & 29.6 & 29.0 \\
NN DeID & 22.8 & 24.2 & 20.2 \\

\midrule
\multicolumn{4}{c}{\textbf{$\mathbf{< 5\%}$ Reidentifiable}} \\ 
IDF  & 3.8 & 71.4 & 78.6 \\
IDF (Table-Aware) & 5.0 & 67.3 & 70.4 \\
NN DeID  & 4.4 & 35.9 & 29.5 \\ 

\midrule
\multicolumn{4}{c}{\textbf{$\mathbf{< 1\%}$ Reidentifiable}} \\ 
IDF & 0.0 & 82.2 & 81.1 \\
IDF (Table-Aware)  & 0.1 & 74.7 & 80.2\\
NN DeID &  0.0 & \textbf{43.5} & \textbf{40.0} \\

\bottomrule

\end{tabular}

\caption{Statistics comparing sets of 1000 documents redacted using different methods at various levels of identifiability. Reidentification rate measures the rate at at least one model in our neural-network ensemble can retrieve the correct profile for a redacted document. Information loss is measured as the percentage change in the size of the text when compressed.
}
\label{table:deid}
\end{table*}

\section{Experimental Setup}

\paragraph{Models}
We call our main deidentification model \textit{NN DeID}. We consider several different parameterization variants of the dual encoder.  For the document encoder ($f(x, z)$), we consider two different pretrained language models: RoBERTa-base version \cite{Liu2019RoBERTa} (125M parameters)
and PMLM~\cite{Liao2020Pmlm} (125M parameters), a pretrained encoder specifically designed to support masking-style inference. 
For the table encoder ($g(z)$), we consider: RoBERTa-base  \cite{Liu2019RoBERTa} (125M parameters) with a simple linearized version of the profile, and TAPAS base~\cite{Herzig2020TaPas} (111 million parameters), a model designed to handle table input. 
We randomly compute masks online during training, so documents take a new randomly-redacted form on each epoch.
All models are implemented in Hugging Face transformers library \cite{Wolf2020Transformers}. Each model is trained for sixty epochs, about two days on a single NVIDIA RTX A6000 GPU. More training details are available in \ref{app:training-details}.

We experiment with all combinations for reidentification models, specified by document-profile encoders, RoBERTa-RoBERTa (\textit{RR}), RoBERTa-TAPAS (\textit{RT}), PMLM-RoBERTa (\textit{PR}),  PMLM-TAPAS (\textit{PT}). The PT model is the default for NN DeID.

\paragraph{Baselines}

We consider several unsupervised redaction baselines based on lexical matches with the table and word frequencies.  \textit{Lexical} removes all overlapping words that appear in the profile from the document. 
   \textit{IDF (Table-Aware)} masks all overlapping words that appear in the profile from the document, then masks in order of descending Inverse Document Frequency (IDF) (rarest word first) until a fixed threshold.  We compute IDF based on the full corpus of documents and profiles from the train, validation, and test sets.  \textit{Named entity} removes all named entities from the document.\footnote{We identify named entities using the \texttt{dslim/bert-base-NER-uncased} model available from Hugging Face.
   Named entities identified are personal names (PER), organization names (OR), location names (LOC), and miscellaneous names (MISC) \cite{Tjong2003CoNLLNER}.} 

\paragraph{Metrics}

A major challenge is how to evaluate text privacy in the presence of a strong reidentification models. As shown in Section~\ref{sec:motivate}, information retrieval metrics work well for lightly redacted documents, but fail under heavy masking. We ran preliminary experiments with human subjects, but found that even at seemingly low levels of masking, documents were nearly impossible for humans to reidentify. 

Inspired by work on adversarial privacy such as the NeurIPS Hide-and-Seek challenge \cite{Jordon2021NeurIPSChallenge}, we adapt a metric that utilizes an ensemble of reidentification models $\cal R$ as a benchmark. A masked document, $x, z$, is considered reidentified if any of the models can correctly select its profile, i.e. $\hat{y} = \arg\max_y p_r(y \ | \ x,\  z)$ for any model $r \in {\cal R}$. In order to diversify the ensemble we utilize different pretrained neural models as discussed above. We observe that each model can reidentify others with high accuracy indicating diversity features (more discussion in Section~\ref{sec:diverse}). We also include a word-matching based IR model in the ensemble, but find that it is not competitive at reidentification. Explicitly, the ensemble consists of the three variant parameterizations (\textit{RR, PR, RT}) as well as the IR matching model.  

As a metric of utility, we compute the average percentage of words masked, as well as the information loss percentage, computed as the ratio between the size of the original and redacted texts when compressed. For each method and baseline we sweep over mask sizes to compute a curve of reidentifiability and utility. 

\paragraph{Inference}

\begin{figure}
    \centering
    \includegraphics[width=1.1\linewidth]{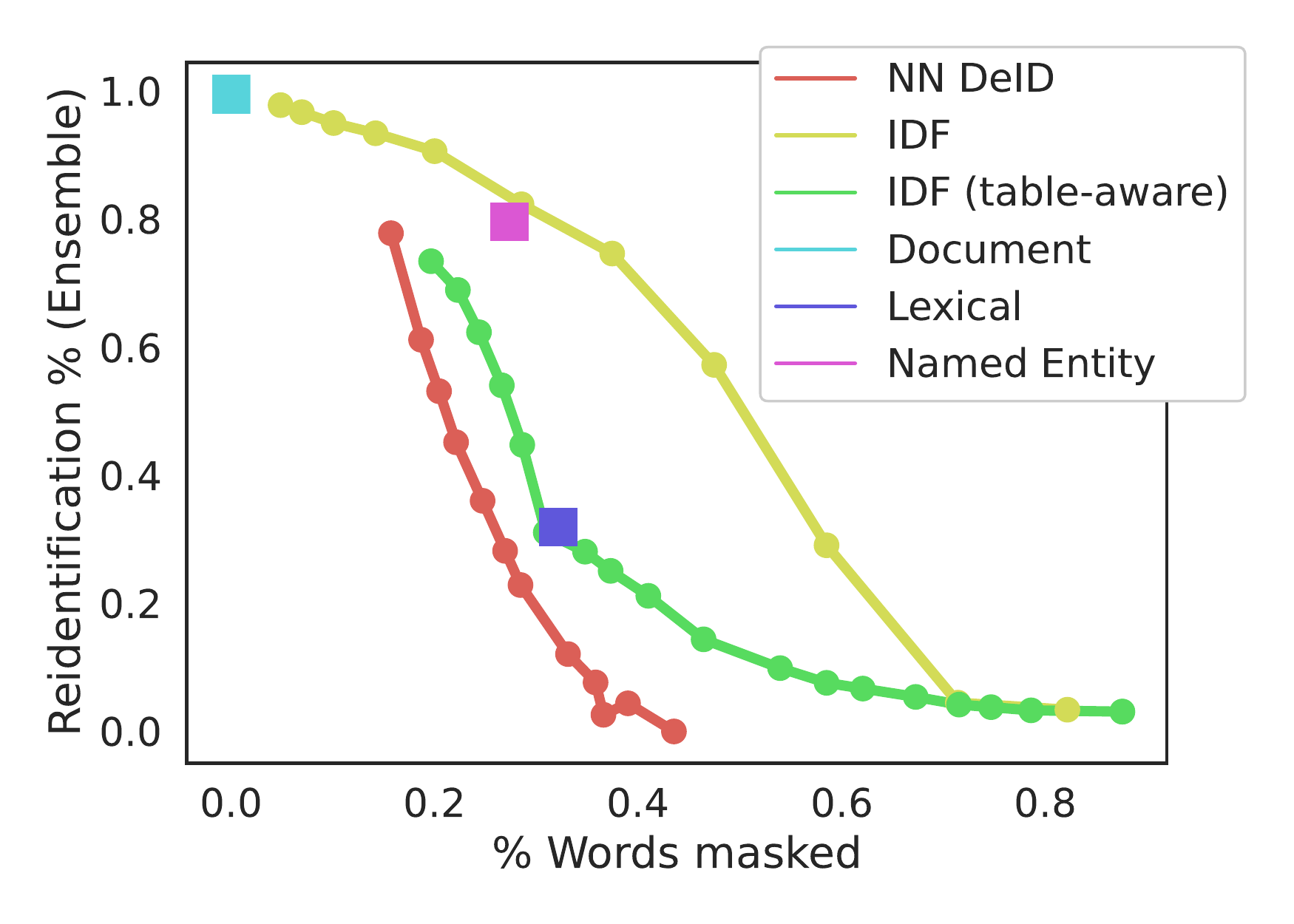}
    \caption{Pareto curves comparing deidentification approaches on privacy versus words masked. Lexical and Named Entity baselines are fixed values. NN DeID is computed with different $K$-anonymity values. IDF (table-aware) and IDF remove until a fixed IDF cutoff threshold.}
    \label{fig:pareto}
\end{figure}

We generate redactions from the reidentification models using greedy search to find the word to mask that causes the maximum decrease in the correct prediction. We use search implementations from the TextAttack library \cite{Morris2020TextAttack}. Search takes in a stopping parameter $K$ which indicates the rank cutoff of $\hat{y}$ to stop search, $\hat{y} \not \in K\arg\max_{y} p(y | x, z)$. We run with different values of $K$ to sweep over levels of privacy, and generate redactions with different masking rates. We ignore stopwords to speed up the search since they will rarely be identifiers.

\section{Results}

\begin{table}[]
\centering
\begin{tabular}{lcc}
\toprule
 &  \multicolumn{2}{c}{Masked}\\
\textbf{Model} & \textbf{0\%} & \textbf{30\%} \\ 
\midrule
Baseline & 52.9 & 10.6 \\
+ Word dropout & 55.4 & 20.5\\
\ \  \ \   Dropout by IDF-weighting & 48.6 & 20.2\\
+ Label smoothing ($\alpha = 0.1$) & 56.3 & 10.8 \\
+ Bigger emb. ($768 \rightarrow 3072$) & 61.8 & 22.2\\ 
+ Table encoder optimization & 98.1 & 14.0 \\ 
\midrule
\textbf{+ Combined} & 96.4 & 38.3 \\ 

\bottomrule
\end{tabular}
\caption{Ablation study. Effect of different factors on model ReID accuracy across data with different redaction strategies. Experiments are on RT parameterization and use 1/10 training data and number of profiles.}
\label{table:model-ablations}
\end{table}

Table~\ref{table:deid} presents results comparing unsupervised deidentification techniques on privacy and utility under the ensemble reidentification metric. As noted above, we see that neither Lexical nor Named Entity redaction provide sufficient privacy. NN DeID can provide better privacy while masking fewer words. Both NN DeID and IDF based approaches can reach stronger levels of privacy ($<5\%$ reidentifiability), but at these levels IDF masks most of the remaining words. 
At full deidentification under the ensemble, NN DeID masks less than half of the words. 
When we consider an information loss measure of utility, NN DeID also performs much better than IDF-based deidentification. 

Figure~\ref{fig:pareto} expands on these results by showing the 
Pareto curves for privacy and utility across methods. Curves are obtained by varying the $K$ value used in NN DeID and the threshold for IDF based deidentification. 
Curves show that in addition to achieving better utility at very low rates of identifiablity, the method also achieves better utility than lexical matching, and a steeper privacy curve even at lower levels of redaction. 

\paragraph{Model ablations}

Table~\ref{table:model-ablations} shows an ablation study of the components added to the model to improve accuracy. 
An alternative approach to this task is to finetune a pretrained model directly for the reidentification task (baseline). However, we found that out-of-the-box this model was neither effective as a pure reidentification model nor as a model to guide search. 
We ablate each component added to NN DeID independently utilizing 1/10th of the training data and profiles, and compare both on the original documents and on documents with 30\% of the words masked.  Word dropout with the proposed sampling rate improves model accuracy particularly in the high mask regime. Interestingly weighting word dropout frequency using IDF hurts model accuracy in the full regime, and is not included in the final model. Increasing the dual encoder embedding sizes from 768 to 3072 and adding label smoothing both increase model accuracy. 
Finally, using coordinate ascent to optimize the profile encoder in addition to the document encoder has by far the largest impact on model accuracy. The combination of these approaches gives a deidentification model that is accurate across levels of masking.

\section{Analysis}

\begin{figure}
    \centering
    \includegraphics[width=0.9\linewidth]{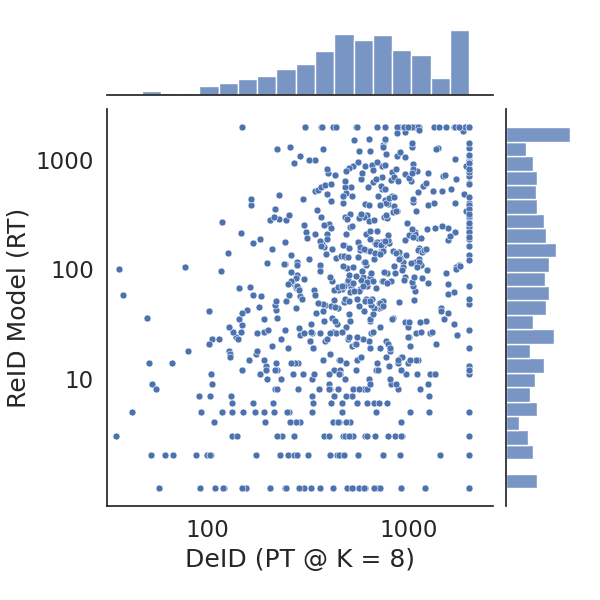}
    \caption{Rank comparison of the true document ($\hat{y}$) in two different parameterized models of $p(y\ |\ x, z)$ (RT and PT). Mask $z$ comes from a deidentification ($K=8$) on the PT model. While correlated, the two parameterizations produce very different rankings.}
    \label{fig:diversity}
\end{figure}

%
\definecolor{mypink}{HTML}{303080}
\definecolor{myblue}{HTML}{303080}

%
\newcommand{\tmaskp}[1]{\textcolor{mypink}{\textbf{\underline{\smash{#1}}}}}
\newcommand{\hmaskp}{\textcolor{mypink}{\sethlcolor{mypink}\hl{\<mask>}}}
\newcommand{\tmaskb}[1]{\textcolor{myblue}{\textbf{\underline{\smash{#1}}}}}
\newcommand{\hmaskb}{\textcolor{myblue}{\sethlcolor{myblue}\hl{\<mask>}}}

\begin{table*}[t]
\scriptsize
\centering
\begin{tabular}{p{0.5\linewidth}p{0.5\linewidth}}
\toprule

Model prediction: \textbf{Dean Roland} (99\%) & Model prediction: \textbf{Avey Tare} (48\%) \\
\tmaskb{Dean} \tmaskb{Roland} (born \tmaskb{October} \tmaskb{10}, \tmaskb{1972}) is \tmaskb{an} \tmaskb{American} musician. He is best known for being the \tmaskb{rhythm} \tmaskb{guitarist} of the \tmaskb{band} Collective \tmaskb{Soul}, an \tmaskb{alternative} \tmaskb{rock} \tmaskb{band} fronted by his \tmaskb{older} \tmaskb{brother} \tmaskb{Ed}. He is also part of the \tmaskb{rock} duo Magnets \& Ghosts alongside \tmaskb{Ryan} \tmaskb{Potesta}.
&
\hmaskb\ \hmaskb\ (born \hmaskb\ \hmaskb, \hmaskb) is \hmaskb\ \hmaskb\ musician. He is best known for being the \hmaskb\ \hmaskb\ of the \hmaskb\ Collective \hmaskb , an \hmaskb\ \hmaskb\ \hmaskb\ fronted by his \hmaskb\ \hmaskb\ \hmaskb. He is also part of the \hmaskb\ duo Magnets \& Ghosts alongside \hmaskb\ \hmaskb.

\\

Model prediction: \textbf{Madoko Hisagae} (100\%) & Model prediction: \textbf{Hiroki Ichigatani} (5\%) \\
\tmaskp{Madoka} \tmaskp{Hisagae} (born 11 January \tmaskp{1979}) is a \tmaskp{Japanese} fencer. \tmaskp{She} competed in the \tmaskp{women's} individual sabre events at the 2006 and 2008 Summer Olympics.
&
\hmaskp\ \hmaskp\ (born 11 January \hmaskp) is a \hmaskp\ fencer. \hmaskp\ competed in the \hmaskp\ individual sabre events at the 2006 and 2008 Summer Olympics.
\\ \\

Model prediction: \textbf{Tim Tolkien} (100\%) & Model prediction: \textbf{Nesbert Mukomberanwa} (6\%)\\

\tmaskp{Tim} \tmaskp{Tolkien} (born \tmaskp{September} \tmaskp{1962} ) is an \tmaskp{English} sculptor who has designed several monumental sculptures, including \tmaskp{the} award - winning \tmaskp{Sentinel}. He has a wood carving and metal \tmaskp{sculpture} business at \tmaskp{Cradley} Heath, West \tmaskp{Midlands}. 
&
\hmaskp\ \hmaskp\ (born \hmaskp\ \hmaskp\  ) is an \hmaskp\  sculptor who has designed several monumental sculptures, including \hmaskp\  award - winning \hmaskp\ . he has a wood carving and metal \hmaskp\  business at \hmaskp\ Heath, West \hmaskp\ .
\\ 
\\
Model prediction: \textbf{Lee Harding (writer)} (97\%) & Model prediction: \textbf{Alan Burridge (writer)} (9\%)
\\
\tmaskb{Lee} \tmaskb{John} \tmaskb{Harding} (born 19 \tmaskb{February} \tmaskb{1937}) is an \tmaskb{Australian} freelance \tmaskb{photographer}, who became a writer of \tmaskb{science} fiction novels and short stories. &
\hmaskb\ \hmaskb\ \hmaskb\ (born 19 \hmaskb\ \hmaskb) is an \hmaskb\ freelance \hmaskb, who became a writer of \hmaskb\ fiction novels and short stories.
\\
\\
Model prediction: \textbf{Begziin Yavuukhulan} (100\%) & Model prediction: \textbf{Tarzi Afshar} ( 25\%) 
\\
\tmaskp{Begziin} \tmaskp{Yavuukhulan}\ (1929 - \tmaskp{1982}) was a \tmaskp{Mongolian} poet of the \tmaskp{communist} era that wrote in \tmaskp{Mongolian} and Russian.
&
\hmaskp\ \hmaskp (1929 - \hmaskp) was a \hmaskp\ poet of the \hmaskp\ era that wrote in \hmaskp\ and Russian.
\\
\\
Model prediction: \textbf{Bob Whiting} (91\%) & Model prediction: \textbf{Bob McDonald} (9\%) \\ 
Robert ``Bob'' \tmaskb{Whiting} (\tmaskb{6} \tmaskb{January} \tmaskb{1883} -- \tmaskb{1917}) was an English footballer who played in the football league for \tmaskb{Chelsea}. \tmaskb{Whiting} died in France whilst fighting in World War \tmaskb{I}. He is commemorated at the Arras Memorial.
&
Robert ``Bob'' \hmaskb\ (\hmaskb\ \hmaskb\ \hmaskb\ -- \hmaskb) was an English footballer who played in the football league for \hmaskb. \hmaskb\ died in France whilst fighting in World War \hmaskb. He is commemorated at the Arras Memorial.
\\
\\

Model prediction: \textbf{Ronald Jonker} (99\%) & Model prediction: \textbf{Peter McDermott} (94\%) \\
\tmaskb{Ronald} \tmaskb{Jonker} (born \tmaskb{14} December 1944) is a former Australian cyclist. He competed in the individual road race at the 1968 Summer Olympics.
&
\hmaskb\ \hmaskb\ (born \hmaskb\ December 1944) is a former Australian cyclist. He competed in the individual road race at the 1968 Summer Olympics. \\
&
\\
Model prediction: \textbf{Brad Turner} (93\%) & Model prediction: \textbf{Andrew Dosunmu} (10\%)
\\
\tmaskp{Brad} \tmaskp{Turner} is a Canadian film director, television director, and photographer.
&
\hmaskp\ \hmaskp\ is a Canadian film director, television director, and photographer.
\\
\\
Model prediction: \textbf{Julie Roginsky} (93\%) & Model prediction: \textbf{Ann Curry} (11\%) \\

\tmaskb{Julie} \tmaskb{Roginsky} (born \tmaskb{April} 25, \tmaskb{1973}) is a Democratic Party \tmaskb{strategist} and television personality. She is a contributor with the Fox news channel and a co-host of The Five. (…) &  
\hmaskb\ \hmaskb\ (born \hmaskb\ 25, \hmaskb) is a Democratic Party \hmaskb\ and television personality. She is a contributor with the Fox news channel and a co-host of The Five. (…) \\



\bottomrule
\end{tabular}
\caption{Example redactions from the system.}
\label{table:redaction-examples}
\end{table*}

\begin{figure}
    \centering
    \vspace{0.2cm}
    \includegraphics[width=0.9\linewidth]{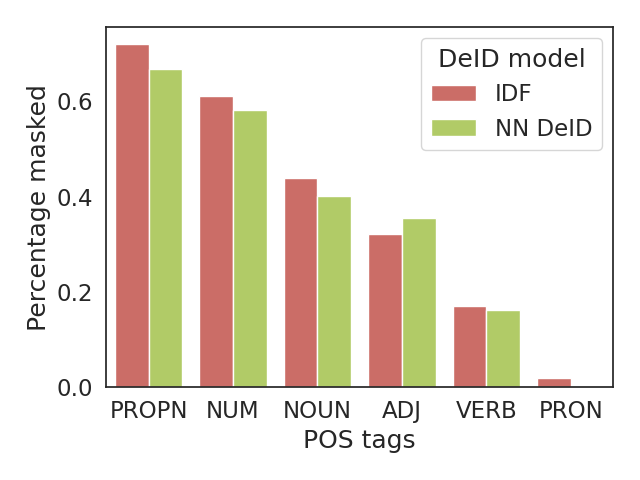}
    \caption{Percentage of words by part-of-speech tags that are masked by the IDF model and NN ReID model at $K=8$ (similar masking level).  }
    \label{fig:pos}
\end{figure}

\subsection{Quasi-Identifiers in Redacted Examples}

Table \ref{table:redaction-examples} shows examples of redacted documents. While the most common redacted entities in deidentified examples are names, dates, and locations, we  find notable examples of redacted quasi-identifiers:
\begin{itemize}

    \item  Determiners. Determiners can provide useful information in context.  In the first example, the system removes ``American'' before musician, but also the word ``an'' which, in this context, signals the next word may be ``American''. This example is also interesting in that it preserves the word ``Collective'', leading the model to predict a musician Avey Tare from the band ``Animal Collective". 
    \item Gender markers.  The model often redacts words marking gender in order to anonymize documents. In the second example, for the document on Madoko Hisagae, the model removes both the word ``She" and ``women's". This redaction leads to the prediction of Hiroki Ichigatani as the predicted match, a male Olympic fencer. 
    \item Locations. The pretrained model seems to be able to identify relative locations even if they are not represented directly in the profile. In the third example, the profile indicates that Tim Tolkien is an English sculptor. The word ``English'' is masked immediately, but the location ``Cradley Heath, West Midlands'' is a quasi-identifier as to the country. Upon redacting this term, the model switches its prediction to Nesbert Mukomberanwa, a sculptor from Zimbabwe. 

\end{itemize}

\definecolor{myred}{HTML}{803030}

%
\newcommand{\hmaskw}{\textcolor{myred}{\sethlcolor{myred}\hl{\<mask>}}}

\begin{table*}[t]
\scriptsize
\centering
\begin{tabular}{p{0.5\linewidth}p{0.5\linewidth}}
\toprule

Model prediction: \textbf{J.G. Blackman} (99\%) & Model prediction: \textbf{J.G. Blackman} (28\%) \\
J. G. Blackman was a West Indian cricket umpire. He stood in one test match, West Indies vs. England, in 1935 . 
&
\hmaskw. \hmaskw. \hmaskw\ \hmaskw\ \hmaskw\ \hmaskw\ \hmaskw\ \hmaskw\ \hmaskw. \hmaskw\ \hmaskw\ \hmaskw\ \hmaskw\ \hmaskw\ \hmaskw\ , \hmaskw\ \hmaskw\ \hmaskw. England, \hmaskw\ \hmaskw. 
\\
\\
Model prediction: \textbf{Nadezhda Shitikova} (99\%) & Model prediction: \textbf{Nadezhda Shitikova} (11\%) \\
Nadezhda Shitikova ( ; 15 September 1923 -- 1995 ) was a Soviet fencer. She competed in the women's individual foil event at the 1952 and 1956 Summer Olympics.
&
\hmaskw\ \hmaskw\ ( ; \hmaskw\ \hmaskw\ \hmaskw\ -- \hmaskw\ ) \hmaskw\ \hmaskw\ Soviet \hmaskw. \hmaskw\ \hmaskw\ \hmaskw\ \hmaskw\ \hmaskw'\hmaskw\ \hmaskw\ foil \hmaskw\ \hmaskw\ \hmaskw\ \hmaskw\ \hmaskw\ \hmaskw\ \hmaskw\ \hmaskw.
\\
\\
Model prediction: \textbf{Begziin Yavuukhulan} (98\%) & Model prediction: \textbf{Begziin Yavuukhulan} (9\%) \\
Begziin Yavuukhulan ( , 1929-1982) was a Mongolian poet of the communist era that wrote in Mongolian and Russian.
&
\hmaskw\ \hmaskw\ ( , 1929-1982) \hmaskw\ \hmaskw\ \hmaskw\ \hmaskw\ \hmaskw\ \hmaskw\ \hmaskw\ \hmaskw\ \hmaskw\ \hmaskw\ \hmaskw\ \hmaskw\ \hmaskw\ \hmaskw.
\\
\\
Model prediction: \textbf{Sally Raguib} (100\%) & Model prediction: \textbf{Sally Raguib} (31\%) \\
Sally Raguib (born 8 September 1996) is a Djiboutian Judoka. She competed in the women's 57 kg event at the 2012 Summer Olympics.
&
\hmaskw\ \hmaskw\ (\hmaskw\ 8 \hmaskw\ 1996) \hmaskw\ \hmaskw\ \hmaskw\ \hmaskw. \hmaskw\ \hmaskw\ \hmaskw\ \hmaskw\ \hmaskw'\hmaskw\ \hmaskw\ \hmaskw\ \hmaskw\ \hmaskw\ \hmaskw\ \hmaskw\ \hmaskw\ \hmaskw.
\\
\bottomrule
\end{tabular}
\caption{Examples of redactions where our neural ensemble can correctly reidentify the individual at extremely high levels of document masking, even though the documents were never seen during training.}
\label{table:redaction-examples-high-masking}
\end{table*}

\subsection{Redacted Word Types}

The IDF (table-aware) model relies on overlapping words and rare words to redact content, whereas the NN DeID model can in theory remove any identifying word.  Figure~\ref{fig:pos} compares the part-of-speech tags of the masked words between the two models at the same redaction level. We see that while similar, the NN DeID model masks fewer nouns, proper nouns and numbers, and more adjectives and pronouns. These word classes are less likely to fit the IDF or table-matching criterion.

\subsection{Model Diversity}
\label{sec:diverse}

The ensemble used for deidentification contains three separate pretrained encoder variants. One potential issue is that the model used to deidentify the text may be overly correlated with the ensemble models used for evaluation. However, we find that each model is quite strong on reidentifying redactions made by other models. For example, the RR model can reidentify NN DeID (PT, K=1) with a surprisingly high 60.5\% accuracy. In general we find the model rankings are quite different.

Figure~\ref{fig:diversity} demonstrates this phenomenon. In this figure, examples are deidentified to $K=8$ with a PT parameterization, and we plot a rank-rank joint histogram with an RT parameterization. While there is some correlation in the rankings, the two models produce very different rankings, with RT even fully reidentifying some points.

\subsection{Reidentification at high levels of masking}
\label{sec:reid_mask}
Table \ref{table:redaction-examples-high-masking} shows examples of documents where our reidentification ensemble can correctly identify the individual even at extremely high levels of masking. Examples are randomly generated with a minimum of $95\%$ of words masked. Because we permit punctuation in redacted examples, and we mask but do not erase words, models are able to exploit word counting and punctuation-specific features to identify individuals under very high masking rates.

\section{Conclusion}
We propose an unsupervised method for text deidentification that focuses on deidentifying pseudo-identifiers. 
The method first learns to reidentify from text utilizing a prior masking models. We then utilize search to find a mask to ensure K-anonymity in this model. This approach outperforms masking based on named entities and matching with tabular data, both of which fail to fully anonynize the document. Using an ensemble of reidentification models as a metric, we show that our approach can reach 
high levels of privacy with moderate levels of redaction. In future work we plan to utilize this approach in conjugation with downstream tasks in order to further demonstrate the utility of the redacted data. We also plan to compare and evaluate with  domain-specific approaches for distributing redacted models through manual and automatic redaction.

\section{Limitations}

\paragraph{Issues with Wikipedia.} Many Wikipedia biographical articles within a given category follow a similar syntactic template, so it is possible that a model could learn to partially reidentify a person by looking at superficial features of the article structure. In the future, documents could be paraphrased during training to prevent the model from learning such syntactic idiosyncrasies. Additionally, since RoBERTa and TAPAS's pre-training data both include Wikipedia articles \cite{Liu2019RoBERTa,Herzig2020TaPas} it is possible that the models can ``cheat'' on the test set by recalling data that they memorized during their pre-training. We hypothesize that cheating is unlikely to be happening for two reasons. First, articles in Wikibio make up a small percentage of the models' training data, so very little of their information is probably stored in the pre-trained weights of the models. Second, the models' performance on the test set before training is very low (0\% test accuracy). Finally, Wikibio contains articles about a very small and biased subset of humanity \cite{Yuan2021SynthBio}.

\paragraph{Need for a profile.} Although the method we propose does not require any labeled data, it requires a different new data source in the form of profiles. This means that the information deidentified is limited to what can be captured in the profile. Thus, the work of adapting this to a new domain shifts from collecting human-labeled PII annotations to collecting as much personal information as possible into profiles. This is much easier in domains like medicine where a great deal of personal information is known about each patient, but collecting such profiles may not be possible in every scenario.

\paragraph{Number of words as a quasi-identifier} This work focuses on redacting data by replacing words with masks. One unaddressed issue in this work is the fact that even when masked, the presence of a word can still leak information. Consider the following example: ``Jack Leswick (January 1, 1910 -- August 4, 1934) was a Canadian ice hockey centre for the \texttt{<mask> <mask> <mask>}.''. Leswick's team, the Chicago Black Hawks, is one of 11 of 32 National Hockey League teams with three words in their name. An adversary can eliminate the possibility that Leswick played for any of the 20 two-name teams. Future work can consider the possibility of deleting words entirely or joining multiple masked words into a single mask token to provide additional privacy.

\paragraph{Hiding in the crowd.} K-anonymity exists when an individual cannot be distinguished from $K-1$ other individuals in the dataset. This means that for a given individual, all anonymity guarantees in our setting are with respect to the other individuals in the dataset. Therefore, the same document could be deidentified differently depending on which other profiles there are in the dataset, even without any changes to the document itself.


\section{Ethical Considerations}

This paper targets deidentification, a technique which has been used to democratize access to sensitive data in business, law, and healthcare. However, this paper also discusses the topic of reidentification, and raises issues about how models that identify individuals from seemingly-anonymized data may be used in a negative manner. Reidentification models may be used as part of linkage attacks, where individuals can be pinpointed even from seemingly anonymized data. Additionally, the world knowledge of today's large language models may be well-suited for this type of linkage attack. We observed this behavior empirically, when our models were uncannily able to reidentify individuals within a dataset of $720,000$ identities, even from documents that appeared to have no remaining personal information. 

We plan to release our models for deidentifying documents from Wikibio to the general public. We are open to hearing from users how our technology impacts both their lives and the lives of others, positively or negatively. If we receive any reports of misuse of our technology, we will mitigate accordingly.

\section{Acknowledgments}

AR and JC are supported by NSF CAREER 2037519, NSF 1704834, and a Sloan Fellowship. RZ is supported by a gift from the Simons foundation. JM is supported by Weill Cornell Medicine. Thanks to Dr. Curtis Cole and Dr. Thomas Campion from Weill Cornell Medicine for their general influence on our research direction within the area of deidentification.

\bibliography{acl2022}
\bibliographystyle{acl_natbib}

\appendix

\section{Training details}
\label{app:training-details}

We train all models using the Adam optimizer with an initial learning rate of $5 * 10^{-5}$ or $1 * 10^{-4}$. We clip gradients to a maximum norm of $5.0$. We decrease the learning rate by a factor of $0.5$ whenever performance on a set of held-out redacted validation examples decreases. We train on the full dataset for a duration of 60 epochs, but stop early if the learning rate does not increase for 5 epochs. We implement training using the PyTorch lightning library \cite{FalconPyTorchLightning2019}. We use PMLM-a, the version of PMLM with absolute positional embeddings \cite{Liao2020Pmlm}. All encoders have a maximum sequence length set to 128 throughout all experiments. We truncate tables by dropping columns until the encoded table fits the maximum sequence length.

We use linear learning rate scheduling. For models with the RoBERTa document encoder, we decay the learning rate from a starting point of $1e-5$ to $1e-6$ over the epochs. For models with PMLM as the document encoder, we start the learning rate at $5e-5$. This is because we found empirically the PMLM tended to converge better when started at a lower learning rate. For the first two epochs (one for each encoder), we employ linear warmup, and increase the learning rate from 0 to its true initial value. We find that training the profile encoder is not useful after a handful of epochs, as the profile encoder starts to overfit, and our compute is better spent training the document encoder, which learns much more slowly since its inputs are $50\%$. redacted on average. Thus, after the first 10 epochs (5 of which are spent training the profile encoder), we only train the document encoder.

\section{Search methods ablation}

Our deidentification method redacts words by greedily selecting the word that minimizes the performance with respect to a reidentification model. We also tested using beam search to select words to redact, and found that it did not improve performance. At $k = 1$, beam search with beam width $4$ masked $14.96\%$ of words at $78.1\%$ reidentification rate, while greedy masks $15.46\%$ at a $78.9\%$ reidentification rate, while being $3.39$ times faster. 




\end{document}